%% file: main.tex
\documentclass[sigconf]{acmart}
\AtBeginDocument{%
  }

\setcopyright{acmlicensed}
\copyrightyear{2025}
\acmYear{2025}
\acmDOI{XXXXXXX.XXXXXXX}
\acmConference[HRI '26]{March 16--19,
  2026}{Edinburgh, Scotland}

\usepackage{adjustbox}

\usepackage{multirow}
\usepackage{makecell}

\usepackage[normalem]{ulem}
\usepackage{pifont}
\newcommand{\cmark}{\textcolor{green}{\ding{51}}}%
\newcommand{\xmark}{\textcolor{red}{\ding{55}}}%

\begin{document}


\title{Embodied Referring Expression Comprehension \\in Human-Robot Interaction}

\author{Md Mofijul Islam$^\ddagger$$^\S$$^*$, Alexi Gladstone$^\ddagger$, Sujan Sarker$^\ddagger$, Ganesh Nanduru$^\ddagger$, Md Fahim$^\dagger$, Keyan Du$^\ddagger$, Aman Chadha$^\bullet$$^\S$$^\dagger$, Tariq Iqbal$^\ddagger$\\$^\ddagger$University of Virginia, $^\bullet$Stanford University, $^\dagger$University of Dhaka, $^\S$Amazon GenAI}
\authornote{Work does not relate to position at Amazon.}

\renewcommand{\shortauthors}{Islam et al.}
\newcommand{\ferfe}{Embodied Reference Expression Comprehension}
\newcommand{\erfe}{E-RFE}
\newcommand{\frfe}{Referring Expression Comprehension}
\newcommand{\rfe}{RFE}
\newcommand{\pa}{MuRes}
\newcommand{\rerep}{MuRes}
\newcommand{\ds}{Refer360}


\vspace{0.2in}
\begin{abstract}

As robots enter human workspaces, there is a crucial need for them to comprehend embodied human instructions, enabling intuitive and fluent human-robot interaction (HRI). However, accurate comprehension is challenging due to a lack of large-scale datasets that capture natural embodied interactions in diverse HRI settings. Existing datasets suffer from perspective bias, single-view collection, inadequate coverage of nonverbal gestures, and a predominant focus on indoor environments. To address these issues, we present the Refer360 dataset, a large-scale dataset of embodied verbal and nonverbal interactions collected across diverse viewpoints in both indoor and outdoor settings. Additionally, we introduce MuRes, a multimodal guided residual module designed to improve embodied referring expression comprehension. MuRes acts as an information bottleneck, extracting salient modality-specific signals and reinforcing them into pre-trained representations to form complementary features for downstream tasks. We conduct extensive experiments on four HRI datasets, including the Refer360 dataset, and demonstrate that current multimodal models fail to capture embodied interactions comprehensively; however, augmenting them with MuRes consistently improves performance. These findings establish Refer360 as a valuable benchmark and exhibit the potential of guided residual learning to advance embodied referring expression comprehension in robots operating within human environments.

\end{abstract}

\maketitle

%

\input{sections/introduction}

\input{sections/related_works}

\input{sections/data_collection_system}

\input{sections/annotation_system}
\input{sections/embodied_dataset}
\input{sections/proposed_method}
\input{sections/experiment_setup}
\input{sections/experiment_results}
\input{sections/overal_discussion}

\input{sections/conclusion}

\bibliographystyle{ACM-Reference-Format}
\bibliography{references}

\appendix

\newpage
\input{supplementary/resources}

\input{supplementary/experiments_results}

\input{supplementary/appendix_data_collection_system}

\input{supplementary/appendix-data-annotation-system}
\end{document}

%% file: sections/introduction.tex
\input{sections/dataset_comparison}
\section{Introduction}



\begin{figure}[!t]
    \begin{center}
    \scriptsize
    \includegraphics[width=0.95\columnwidth]{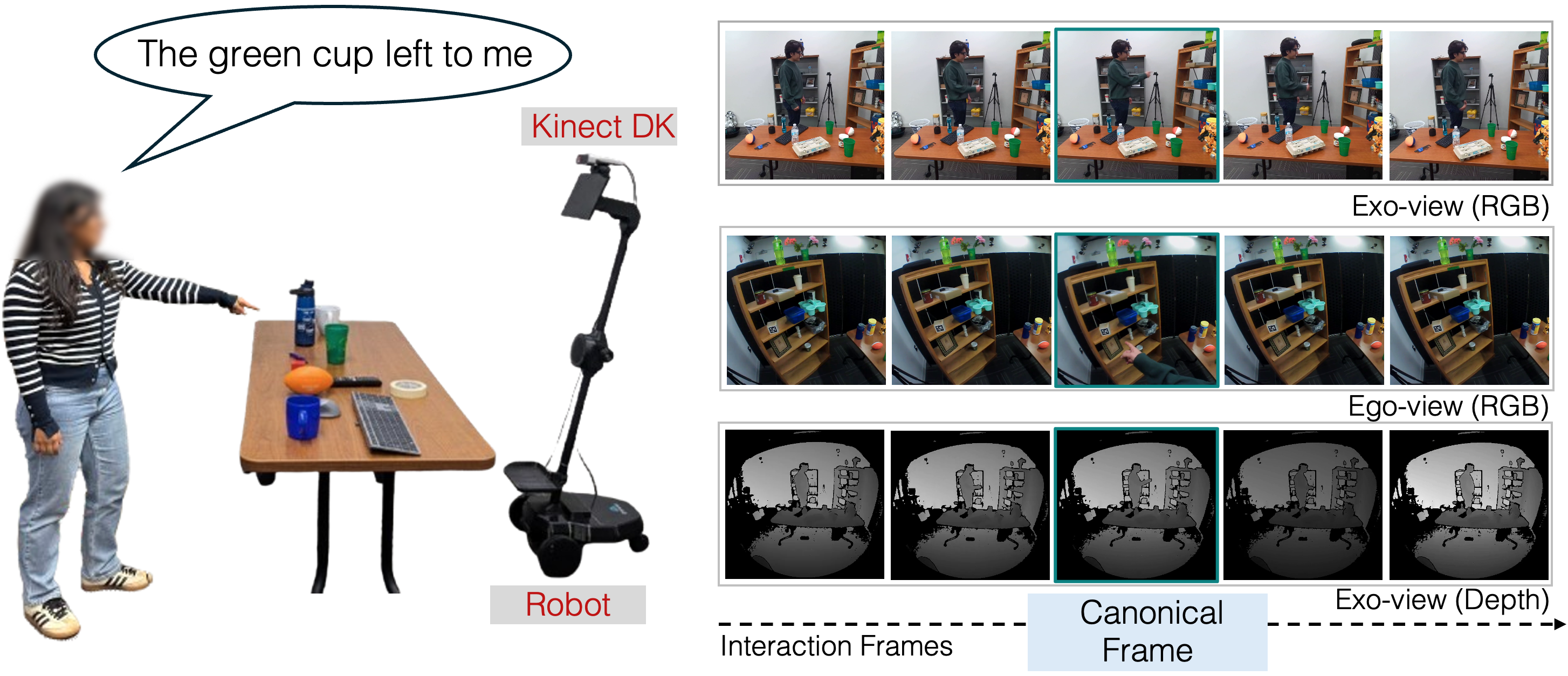}
    \caption{{\ds} data collection setup in human-robot interaction context ({left}). Here, the person is pointing towards an object while verbally describing it. Interaction frames from three different views (exo, ego, and depth). Highlighting the canonical frames, i.e., frames where the subject precisely points to an object ({right}).}
    \Description{Data collection setup in human-robot interaction context ({left}). Here, the person is pointing towards an object while verbally describing it. Interaction frames from three different views (exo, ego, and depth). Highlighting the canonical frames, i.e., frames where the subject precisely points to an object ({right}).}
    \label{fig:data_collection_system}
    \end{center}
\end{figure}


Robots that interact with people, from service robots in public spaces to assistive systems at home and work, must understand instructions given by humans naturally and intuitively \cite{sebo2020robots}. In human–human communication, people utilize multimodal signals to express object reference and intent unambiguously 
\cite{mcneill1992hand,tomasello2010origins}. Similarly, in human–robot interaction (HRI) scenarios, people often communicate their instructions not solely through language; rather, they are grounded in embodied exchanges that combine verbal utterances with nonverbal cues, such as gestures, gaze, and posture \cite{mcneill2012language,arbib2008primate,liszkowski200612,liszkowski2004twelve,tang2020bootstrapping}. For instance, in task scenarios such as passing an object or assembling an item, people rely on both gaze and pointing gestures to specify what ``this one” or ``that side” refers to. Endowing robots with the capability to understand human instructions unambiguously would enable them to be integrated smoothly into human teams and support more natural and intuitive interactions with people \cite{chen2021yourefit,eqa-mx,caesar,kratzer2020mogaze,imprint,yasar2020RAL}.

In the past several years, embodied reference expression (E-RFE) comprehension has received extensive attention in human–robot interaction (HRI), particularly in domains such as assistive interaction and collaborative task execution \cite{stacy2020intuitive,kratzer2020mogaze,islam2020hamlet,islam2021multigat}. The primary goal of much of this work is to enable robots to accurately infer human intent by grounding verbal commands in complementary nonverbal cues such as gaze, gesture, and posture. Such capabilities would allow robots to seamlessly assist humans in assistive scenarios by accurately interpreting instructions and proactively coordinating with human partners in collaborative tasks, thereby improving efficiency and reducing errors. To address this challenge, many early HRI systems often relied on single-modality inputs or highly scripted interactions, which constrained their adaptability and robustness. In contrast, real-world HRI demands robots that can comprehend nuanced, multimodal instructions across dynamic environments and temporal contexts.

While the ability to interpret human instructions is essential for robots, current approaches face significant challenges in real-world HRI scenarios. A primary bottleneck is the scarcity of comprehensive datasets that capture embodied interactions across diverse and natural settings. Existing datasets, such as YouRefIt \cite{chen2021yourefit} and MoGaze \cite{kratzer2020mogaze}, provide real-world embodied interactions, but they have critical limitations that hinder the development of robust comprehension models. First, they include verbal utterances tied to the speaker’s or observer’s perspective (e.g., “left ball” vs. “right ball”), introducing perspective bias that restricts generalization. Second, their reliance on single-view (exo or ego) recordings leads to viewpoint bias, limiting performance in varied environments. Multi-view data (including ego, exo, and top views) is needed to overcome occlusions and capture interaction nuances. Third, they only partially represent nonverbal gestures—typically capturing either pointing or gaze—despite both being complementary for robust comprehension. Finally, most data are collected indoors in constrained settings with stationary cameras, which reduces ecological validity. Together, these limitations prevent existing datasets from supporting models that can fully comprehend embodied interactions in diverse, unconstrained real-world HRI contexts. A comparison of existing datasets is provided in Table~\ref{tab:vqa_eqa_datasets_comparison}.

To address these limitations, we developed a comprehensive and diverse dataset, called Refer360, to facilitate the study of human-robot interactions in real-world settings. We have collected data across both indoor and outdoor environments with varied lighting conditions, object arrangements, and scene appearances (Fig.~\ref{fig:data_collection_system}). Using a multimodal sensing system, we captured interactions from multiple perspectives—including egocentric and exocentric visual viewpoints—as well as depth, skeleton, infrared, audio, gaze, and pupil tracking. Expert annotators annotated all scenes and verbal utterances. The dataset comprises contributions from $66$ participants across $392$ sessions, spanning both laboratory and outside-laboratory environments. In total, the Refer360 dataset contains $13,990$ multimodal interactions, amounting to $3.2$ million synchronized frames ($28,736$ canonical frames) and $17.62$ hours of recording time. 

Beyond dataset biases, another significant challenge in understanding E-RFEs is extracting complementary representations from multimodal data, e.g., verbal instructions and nonverbal gestures. While existing multimodal models fuse multimodal representations from the frozen pre-trained encoders, leading to performance enhancements across various tasks, the representation gap between these frozen representations can result in suboptimal multimodal representations. Several approaches have been proposed in the literature to reduce the representation gap of unimodal signals \cite{alayrac2022flamingo,li2022blip,li2023blip,liu2023visual}. However, fusing these frozen representations using a self-attention or cross-attention approach can overlook modality-specific cues, limiting the model's ability to effectively leverage and integrate the distinct, complementary cues in multimodal interaction signals (verbal and non-verbal). Thus, extracting salient representations across modalities can help to extract complementary representations.


To address this challenge, we introduce a multimodal guided residual module, \rerep, for learning complementary multimodal representations. Unlike existing approaches, MuRes not only aligns cross-modal features but also preserves modality-specific cues through guided residual connections. Inspired by the information bottleneck principle \cite{islam2023representation,wang2022rethinking,tishby2015deep,shwartz2017opening,sun2022graph,alemi2016deep,discrete_bottleneck}, we have designed MuRes as a representation bottleneck that extracts and reinforces the most relevant signals from each modality. This selective reinforcement yields fused representations that are both aligned and modality-aware, enabling a more comprehensive understanding of multimodal embodied interactions. Moreover, MuRes is lightweight and can be seamlessly integrated as an adapter into existing multimodal models, enhancing their representation capacity while remaining practical for robotic implementation.

To evaluate the effectiveness of our module, we conduct extensive experimental analysis on our {\ds} dataset for comprehending referring expressions, alongside three benchmark datasets from the literature. Furthermore, we have integrated {\rerep} into existing multimodal models to show the effectiveness of utilizing {\rerep} for extracting salient complementary multimodal representations. Our experimental analysis suggests that {\rerep} helps improve the performance of these multimodal models. For example, integrating {\rerep} improved the CLIP model's performance (IOU-25) by $3.4\%$ and $4.99\%$ on the {\ds} and CAESAR-PRO \cite{patron} datasets, respectively. In addition to evaluating our method on embodied referring expression (E-RFE) datasets, we also examine its effectiveness on broader visual question answering (VQA) tasks using the ScienceQA \cite{lu2022learn} and A-OKVQA \cite{schwenk2022okvqa} datasets, which reinforced the effectiveness of the proposed method on a broader set of embodied interaction tasks. The results indicated that {\rerep} boosted the VQA accuracy of the VisualBERT model on the ScienceQA \cite{lu2022learn} dataset by $4.58\%$ and the ViLT \cite{kim2021vilt} model on the A-OKVQA \cite{schwenk2022okvqa} dataset by $2.86\%$. These performance improvements depict the significance of our proposed guided residual model for extracting complementary multimodal representations for various downstream tasks. The outcomes of these experiments promise to advance multimodal instruction comprehension by providing a comprehensive dataset and valuable insights for improving state-of-the-art representation learning techniques for HRI.

%% file: sections/dataset_comparison.tex
\begin{table*}[!t]
    \small
    \centering
    
    \caption{Comparison of the QA datasets. Existing VQA and EQA datasets do not contain nonverbal gestures (NV), multiple verbal (V) perspectives (MP), contrastive (C), and ambiguous (A) data samples, and outdoor scene data. $^{\ddagger}$Embodied (E) interactions refer to humans interacting using multimodal expressions.  $^{\dagger}$Embodied interactions refer to an agent navigating in an environment. $^{\star}$Sythetic Environment. \textbf{Please check the supplementary for a detailed comparison with other related datasets.}}

    \begin{tabular}{lccccccccccccc}
        \toprule
          \multirow{2}{*}{\makecell{Datasets}} &  \multirow{2}{*}{\makecell{V}} &  \multirow{2}{*}{\makecell{NV}} &  \multirow{2}{*}{\makecell{E}} &  \multirow{2}{*}{\makecell{MP}} & \multicolumn{2}{c}{\makecell{Views}} & \multirow{2}{*}{\makecell{C}} &  \multirow{2}{*}{\makecell{A}} &  \multirow{2}{*}{\makecell{Image\\Frames}} &  \multirow{2}{*}{\makecell{Interaction\\Samples}}  & \multirow{2}{*}{\makecell{Environment}} &  \multirow{2}{*}{\makecell{Type}} \\ \cmidrule{6-7}
         & & & & &  Exo &  Ego & & & \\ \midrule

         VQA \cite{vqa} &  \cmark &  \xmark &  \xmark &  \xmark &  \cmark &  \xmark & \xmark &  \xmark &  204K &  614K & Internet & Image \\
        
        
        
        
        

         GRiD-3D$^{\star}$ \cite{lee2022right} &  \cmark &  \xmark &  \xmark & \xmark &  \cmark &  \xmark & \xmark &  \xmark &  8K &  445K & Simulated & Image \\

        
         EQA$^{\dagger}$ \cite{das2018embodied} &  \cmark &  \xmark &  \cmark$^{\dagger}$  &  \xmark &  \xmark &  \cmark$^{\dagger}$ &  \xmark &  \xmark &  5K &  5K & Simulated & Interactive \\
        
         MT-EQA$^{\dagger}$ \cite{yu2019multi} &  \cmark &  \xmark &  \cmark$^{\dagger}$ &  \xmark &  \xmark &  \cmark$^{\dagger}$ &  \xmark &  \xmark &  19K &  19K & Simulated & Interactive  \\
        
        
         {CAESAR}$^{\ddagger\star}$\cite{caesar}  &  \cmark &  \cmark &  \cmark & \cmark &  \cmark &  \cmark & \cmark &  \cmark &  841K &   1M & Simulated & Image\\ 

         {EQA-MX}$^{\ddagger\star}$\cite{eqa-mx} &  \cmark & \cmark  &  \cmark &  \cmark &  \cmark &  \cmark & \cmark &  \cmark &  750K &   8K & Simulated & Image\\

         YouRefIt \cite{chen2021yourefit} &  \cmark &  \cmark &  \cmark & \xmark & \xmark &  \cmark &  \xmark & \xmark &  497K &  4K & Indoor & Video \\
        
        \midrule
        
         {\ds}$^{\ddagger}$ &  \cmark &  \cmark  &  \cmark &  \cmark &  \cmark &  \cmark & \xmark &  \cmark &  1.3M &   14K & Indoor+Outdoor & Video\\
        \bottomrule
    \end{tabular}

    \label{tab:vqa_eqa_datasets_comparison}
\end{table*}

%% file: sections/related_works.tex
\section{Related Work}

\subsection{Embodied Referring Expression Datasets} 
In the literature, embodied interactions are studied in two forms. The first involves agents navigating an environment to gather visual data following verbal instructions \cite{das2018embodied,yu2019multi}. The second focuses on comprehending referring expressions involving verbal and nonverbal cues, where agents interpret and respond to human instructions \cite{chen2021yourefit,patron,caesar,eyiokur2025capeclipawarepointingensemble,ZHANG2024108493,kurita2023refegoreferringexpressioncomprehension,tang2024groundinglanguagemultiperspectivereferential}. We explore the second aspect of embodied interactions, focusing on understanding multimodal referring expressions.

Several datasets have been developed in the literature to study embodied referring expressions ({\erfe}). For example, \citet{chen2021yourefit} developed an embodied referring expressions dataset where a human refers to an object using verbal and pointing gestures. In their proposed dataset, \citet{kratzer2020mogaze} focused on capturing the human body motion and eye gaze. To incorporate both verbal and nonverbal signals, \citet{caesar} developed a synthetic dataset by generating nonverbal cues (pointing gesture and gaze) in a virtual environment and template-based verbal instructions. While these datasets demonstrated the importance of developing diverse datasets towards comprehensively understanding {\erfe}, they predominantly focus on indoor settings \cite{chen2021yourefit}, static camera views without agent or human motion \cite{kratzer2020mogaze}, scripted human interactions \cite{patron}, limited sensor modalities \cite{chen2021yourefit,kratzer2020mogaze}, and synthetic environments \cite{caesar,eqa-mx,patron}. Recent HRI work has also highlighted that datasets constrained to narrow scenarios cannot capture the richness of situated human instructions, particularly when gestures, perspective-taking, and environmental diversity are involved \cite{shrestha2024natsgd,thomason2019improving, kollakidou2022hri,higger2023toward,deichler2023learning,shrinah2024design}. Therefore, these datasets provide limited data samples for developing models for a comprehensive understanding of E-RFE in realistic HRI contexts.  

\subsection{Multimodal Representation Learning}  
There has been significant progress in the last several years on developing multimodal models, particularly focusing on Visual Question Answering (VQA) tasks \cite{Lin_2024_CVPR,lu2019vilbert,yang2023unimo3multigranularityinteractionvisionlanguage,hao2024ademvladaptiveembeddedfusion,guo2025mmrlmultimodalrepresentationlearning,li2023blip,kim2021vilt,clip,li2022blip,zhai2022lit,alayrac2022flamingo,liu2023visual,goyal2017making,gao2015you,yu2015visual,zhu2016visual7w,krishna2017visual}. VQA tasks also involve E-REF tasks, but encompasses a more wider range of task families. For example, \citet{li2019visualbert} used a Transformer with Self-Attention to extract salient multimodal representations, which were trained using visually grounded language objectives. ViLT \cite{kim2021vilt} processed visual inputs holistically, learning vision–language representations without relying on the regional supervision typically associated with object detection. \citet{li2023blip} designed a Querying Transformer to bootstrap vision–language representation from a frozen image encoder. These models achieved performance improvements on VQA tasks by utilizing representation alignment-based training objectives.  

However, as these objectives primarily focus on alignment, they are less effective at fusing modality-specific representations. Self- and cross-attention mechanisms mainly align tokens across modalities but often fail to extract complementary signals that are crucial for embodied referring expression comprehension. In HRI, where subtle nonverbal cues such as gaze and gesture significantly alter instruction meaning, the ability to capture complementary features is critical. Recent HRI research has begun to adapt multimodal transformers to embodied settings, showing that fusion modules or adapters improve robots’ ability to follow human instructions and interpret multimodal input in collaborative tasks \cite{shridhar2020alfred, liu2022instruction,jiang2022vima,driess2023palm}. These works highlight the gap between generic VQA-style representation learning and the requirements of embodied instruction comprehension in HRI, motivating the need for approaches that explicitly enhance modality-specific fusion.

%% file: sections/data_collection_system.tex
\begin{figure}
     \begin{center}
     \includegraphics[width=0.95\columnwidth]{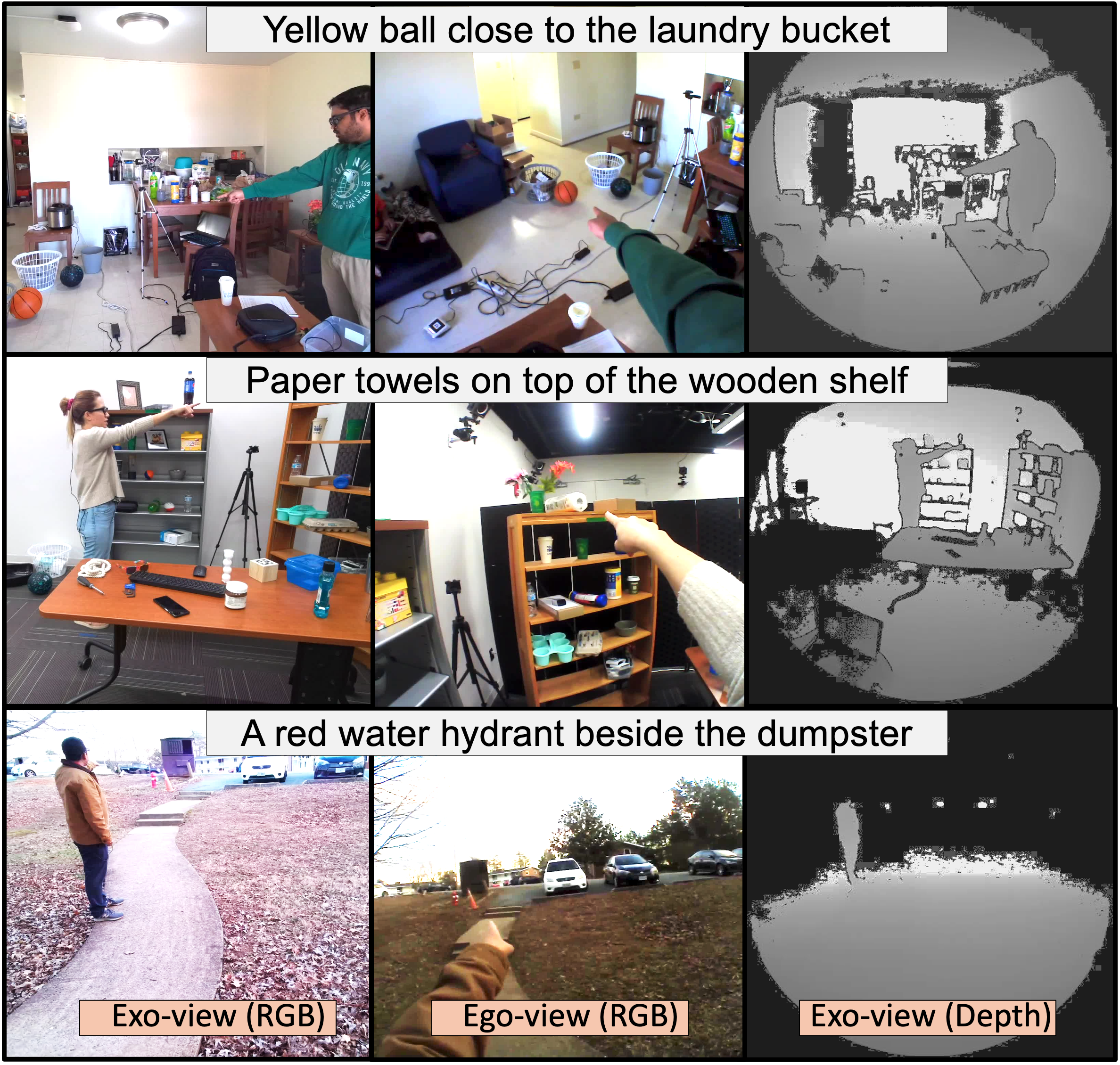}
     \caption{Sample canonical frames from {\ds} dataset in three different views: Exo-view (RGB), Ego-view (RGB), and Exo-View (Depth). The first, second, and third rows contain interaction samples from a home, lab, and outdoor location.}
     \Description{Sample canonical frames from {\ds} dataset in three different views: Exo-view (RGB), Ego-view (RGB), and Exo-View (Depth). The first, second, and third rows contain interaction samples from a home, lab, and outdoor location.}
     \label{fig:diff_view}
     \end{center}
 \end{figure}

\vspace{-.2cm}
\section{Refer360: E-RFE Dataset}

In this section, we introduce Refer360: a multimodal embodied referring expression dataset. Unlike prior datasets limited to controlled settings or single viewpoints, Refer360 provides synchronized recordings across diverse real-world contexts, including both laboratory and unconstrained outdoor environments. As illustrated in Fig.~\ref{fig:diff_view} and summarized in Table~\ref{tab:dataset_statistics}, the dataset integrates complementary sensing modalities: egocentric video and gaze captured through the Pupil Invisible eye tracker, exocentric RGB, depth, infrared, and audio streams from the Azure Kinect DK mounted on an Ohmni telepresence robot \cite{ohmnilabs}, as well as 3D skeletal joint data obtained through Kinect’s body tracking SDK. This configuration yields aligned first-person and third-person perspectives, enriched with gaze and gesture signals crucial for resolving referential ambiguity. By offering a multimodal, multi-view record of natural human instructions, Refer360 can be leveraged for tasks such as perspective-invariant reference resolution, gesture–gaze fusion modeling, and training anticipatory models for HRI.

%
%

\begin{table*}[!t]
   %
    \centering
\caption{Statistical breakdown of {\ds} dataset.}

\begin{tabular}{lcccccc}
\toprule
 & {Sessions}  & { Interactions}  & {Frames}  & Canonical Frames   & {Avg. Interaction Duration}  & {Total Duration}  \\
 \midrule
Lab & 198   & 10,814 & 2,472,939 & 22,356 & 4.484 sec & 13.48 hr \\
Outside-lab & 194   &  3,176 & 759,018 & 6,380 & 4.691 sec & 4.14 hr \\
\midrule
\textbf{Total} & \textbf{392}   & \textbf{13,990} & \textbf{3.2M} & \textbf{28,736} & \textbf{4.531 sec} & \textbf{17.62 hr} \\
\bottomrule
\end{tabular}

\label{tab:dataset_statistics}
\end{table*}

\vspace{-0.02in}
\subsection{Data Collection System}

The goal of the Refer360 dataset is to study real-world HRI in which a human provides object-referencing instructions to robots across diverse environments, spanning controlled laboratory setups to outdoor locations. To achieve this, we have developed a data collection system that synchronously captures multimodal data of embodied interactions in lab and outside-lab environments, utilizing an Azure Kinect DK \cite{azure-kinect} and a Pupil Glass eye tracker \cite{pupil-labs}. It is worth noting that by ``outside-lab environment,'' we encompass settings, including home, outdoor locations, etc.

Figure \ref{fig:data_collection_system} depicts a sample data collection setup of {\ds}. The Azure Kinect DK is mounted on an Ohmni telepresence robot \cite{ohmnilabs} to incorporate camera motion and replicate real-world settings. The Kinect sensor offers multiple data streams that capture different interaction modalities. Its RGB camera continuously records visual data, providing an external or exocentric perspective of the participant's actions. The Pupil eye tracker records an RGB data stream, capturing the participant's first-person or egocentric perspective. Additionally, the Kinect sensor captures depth, infrared, and audio data streams, enabling analysis of the participant's environment and audio cues. We utilize Kinect's Body Tracking SDK to capture 3D skeletal data with 32 body joints, allowing us to track the participant's movements and postures. By combining exocentric and egocentric viewpoints, along with multimodal data from the same interaction, our system offers a comprehensive understanding of embodied human-robot interactions.

In addition to enabling multimodal sensing, incorporating the Ohmni telepresence robot into our data collection system was a critical design choice. During the data collection process, the robot was teleoperated by a researcher. Mounting the Azure Kinect DK on a mobile robotic platform provided two key advantages. First, it allowed us to capture data from the robot’s own perspective, offering a viewpoint that an autonomous system would rely on during interaction. This perspective is particularly important for embodied referring expression (E-RFE) tasks, where the robot must interpret human verbal and nonverbal instructions in real time. Second, teleoperating the robot during experiments provided us with the flexibility to adjust its position and orientation, enabling naturalistic scenarios in which robots dynamically change their viewpoint to enhance understanding. By situating the human instructions in the visible presence of a robot, our setup ensured that the collected data reflected authentic HRI scenarios rather than isolated human behavior, which ensures the ecological validity of the dataset.

We have developed a Python-based application to synchronize the data collection process. It utilizes the pyKinectAzure library for the Kinect sensor's data streams and Pupil Labs' Real-time Python API for the Eye Tracker's data streams. We log the UNIX timestamps of data capture events for multiple sensor data streams from Kinect and Eye Tracker. We used these timestamps to synchronize the captured data during post-processing. This timestamp-based synchronization method can be extended to seamlessly integrate various additional sensors for enhanced functionality and versatility. We will opensource this data collection system for future research. \textbf{Details of the data collection system can be found in Appendix A.}

%
%

%
\subsection{Participants}
\vspace{-.1cm}
After receiving approval from the Institutional Review Board (IRB) for our study involving human participants, we recruited $66$ participants $(n = 66)$ for the study and data collection with $53\%$ males $(n = 35)$ and $47\%$ females $(n = 31)$. The participants were primarily students from various academic backgrounds. The average age of the participants was $26.66$ years, with a standard deviation of 3.36 years. One participant did not consent to release the data. We excluded that participant data from {\ds}. Each participant was compensated $\$15$ for $1$ hour of their time, which is higher than the state minimum wage guideline.
%


\section{Data Collection Procedure}

All data collection tasks involved participants providing object-referencing instructions across multiple sessions, where the environment setup, objects, and viewpoints varied. Our goal was to capture embodied human instructions using both verbal and nonverbal modalities under diverse real-world conditions.

\noindent\textbf{Pre-Task Setup:} Before beginning the study, participants reviewed the consent documents and task instructions, which the University’s Institutional Review Board (IRB) had approved. They then completed a demographic survey, reporting background information such as age, gender, and prior experience with robots. Each participant then wore the Pupil Invisible eye tracker, which provided egocentric video and gaze data, and was introduced to the Azure Kinect DK sensor mounted on an Ohmni telepresence robot, which captured exocentric RGB, depth, infrared, and skeletal joint data. This combination of sensors ensured synchronized multimodal recordings from both first-person and third-person perspectives.

\noindent\textbf{Data Collection Sessions:} Each session lasted approximately one hour and required participants to issue embodied instructions referencing physical objects in their surroundings. The sessions were conducted under two conditions:
\begin{itemize}
    \item \textbf{Constrained condition:} Participants were encouraged to follow guidelines on instruction format and employ both verbal and nonverbal modalities (e.g., gaze, pointing) to reference objects.
    \item \textbf{Unconstrained condition:} No specific guidance on instruction format or modality was given, allowing participants to provide instructions naturally and flexibly.
\end{itemize}

During each interaction, participants were instructed to describe a target object so that a robot could identify it. Instructions could be given from different perspectives (e.g., “the box to my right” vs. “the box to your left”), capturing the variability and ambiguity present in situated communication. To facilitate synchronization, canonical events—such as when a participant pointed at an object or fixated gaze—were timestamped using a keystroke-based system, allowing alignment across all data streams. Objects varied across sessions and included a wide range of everyday household and office items (see Appendix).

\noindent\textbf{Post-Task Survey:} Upon completing the sessions, participants filled out a post-task survey indicating their preferred referencing strategy: verbal-only, gesture-only, or a combination of modalities. They also provided feedback on the naturalness and effectiveness of their interaction style. Finally, participants signed a release form permitting the collected data to be shared as part of the Refer360 dataset. Please refer to the Appendix for further details on the data collection protocol and procedure. 

\subsection{Dataset Post-processing}
%
We have recorded a single video file utilizing the Kinect sensor for each session, which contains three data streams: RGB, Depth, and Infrared. Using the data collection application, we read the Kinect sensor's IMU and 3D skeleton joint data and stored them in separate JSON files. We utilize the FFmpeg library to split the Kinect video stream into three separate streams for RGB, Depth, and Infrared. The IMU time series data is split into two files: accelerometer readings and gyroscope readings. We extracted the recorded audio from Kinect as an MP3 file. For each session, the Pupil eye tracker generates a video file in MP4 format and saves it to the Pupil Cloud with event timestamps. 

One of the major challenges in data post-processing was synchronizing the Azure Kinect and Pupil Eye Tracker data and segmenting each interaction. We used the start and end times of each interaction for segmentation from the Pupil Cloud event timestamps log. Additionally, we logged canonical frames (Figure \ref{fig:data_collection_system} (right)), i.e., frames where participants precisely pointed to the object of interest during data collection. We leveraged the FFmpeg library to split the data into individual interactions and these specific canonical frames for Kinect and eye-tracking data. We used the Pupil Labs’ Real-time Python API for the eye tracker to access the corresponding recordings stored in the Pupil cloud, matching them to the Kinect data using timestamps. Finally, we employed the OpenAI Whisper library to transcribe the audio data captured by the Kinect. Under the approved IRB, five human experts validated all interaction segmentation, synchronization, and audio transcriptions to ensure high-quality data. This dataset was annotated by human annotators from an external company, which provides data annotation services. Figure~ \ref{fig:diff_view} illustrates sample interactions from {\ds} dataset along with the audio transcription. 

%% file: sections/embodied_dataset.tex
\vspace{-.2cm}
\section{Dataset Analysis}



Table \ref{tab:dataset_statistics} provides a comprehensive statistical overview of the {\ds} dataset. Data collection comprised $392$ sessions conducted across both laboratory and outside-lab environments, yielding a total of $13,990$ interactions over $17.62$ hours of recorded activity. These sessions captured approximately $3.2$ million multimodal frames, including $28,736$ canonical frames aligned to key referencing moments such as pointing or gaze fixation. On average, each session lasted $2.69$ minutes and contained $36.65$ frames, while individual interactions spanned roughly $4.53$ seconds. This balance of short interaction cycles and sustained session diversity underscores the dataset’s suitability for studying embodied communication at both fine-grained and contextual levels.

To complement the recorded interactions, post-task survey responses were analyzed to uncover participants’ preferred strategies for object referencing. The results demonstrate a striking consensus: $96.97\%$ ($n = 63$) of participants favored a multimodal approach, combining verbal instructions with non-verbal gestures such as pointing and gaze. Only $3.03\%$ ($n = 2$) reported relying on verbal instructions alone, and notably, no participant selected non-verbal gestures as a standalone strategy. Beyond this quantitative preference, participants also reflected on the naturalness and effectiveness of different modalities, emphasizing that gestures enriched verbal communication by reducing ambiguity and aiding perspective alignment. These findings highlight an important behavioral insight—humans overwhelmingly default to multimodal referencing when interacting in embodied settings, reinforcing the need for models that can integrate verbal and non-verbal cues seamlessly. 


%

%% file: sections/proposed_method.tex
\section{{\pa}: Multimodal Guided Residual Model}
\label{sec:model}

\begin{figure}
    \centering
    \includegraphics[width=.5\columnwidth]{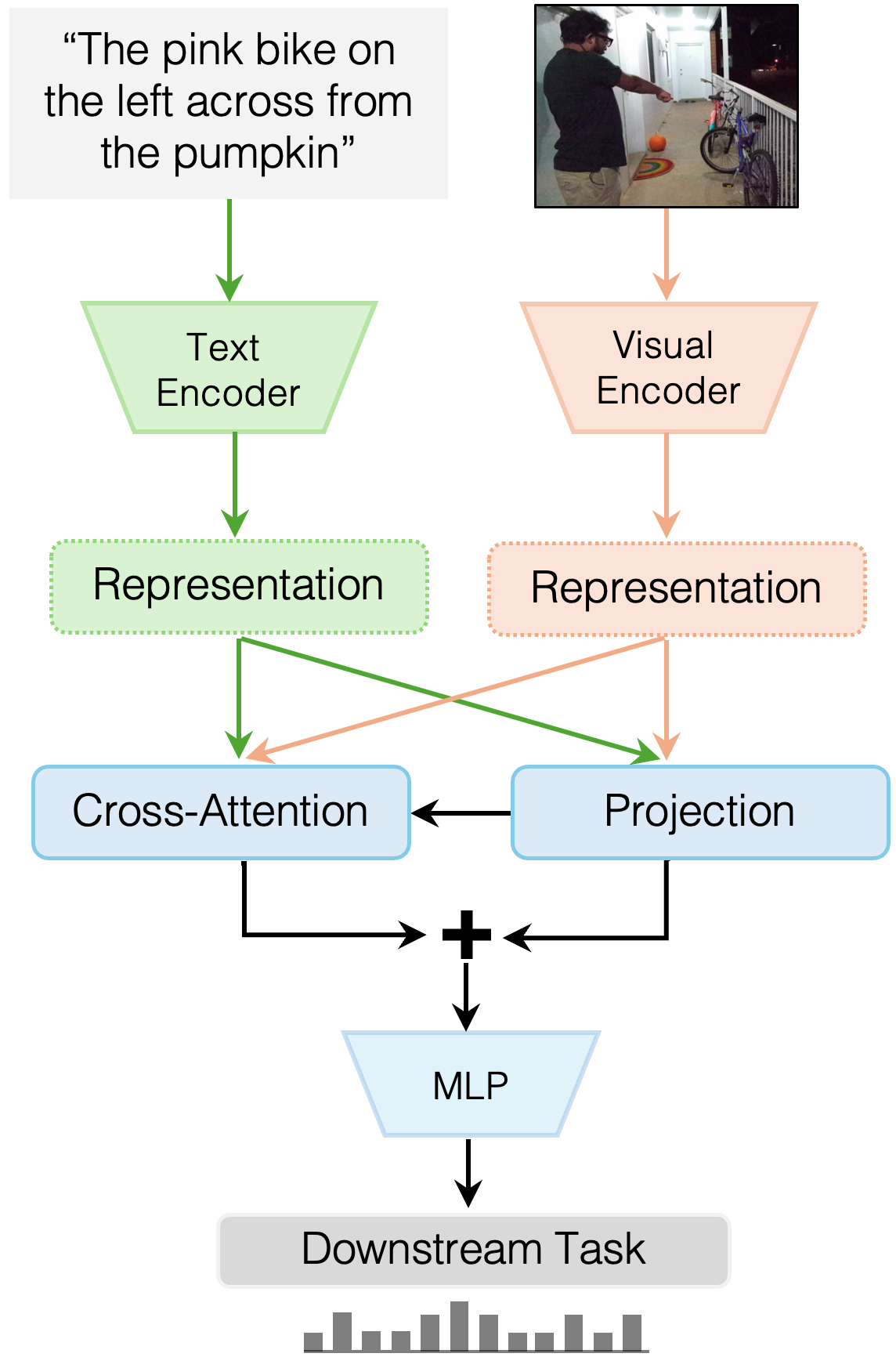}
    \caption{Multimodal Model, {\pa}, with the  Guided Residual module. Visual and language representations are extracted and projected from a pre-trained VL model. The projected representations are fed into the cross-attention module as the query. The key and value are the original extracted visual and language representations on the residual connection. The output from the cross-attention module and the projection are summed for downstream task learning.}
    \Description{Multimodal Model, {\pa}, with the  Guided Residual module. Visual and language representations are extracted and projected from a pre-trained VL model. The projected representations are fed into the cross-attention module as the query. The key and value are the original extracted visual and language representations on the residual connection. The output from the cross-attention module and the projection are summed for downstream task learning.}
    \label{fig:rerep_model_diagram}
    \vspace{-0.15in}
\end{figure}

The task of grounding objects through E-REF requires a comprehensive understanding of verbal utterances and nonverbal gestures. Existing visual-language (VL) models often utilize pre-trained frozen encoders to extract visual and language representations, fusing using self-attention or cross-attention approaches for downstream task learning. These fusion approaches can lose salient information due to the modality gap between frozen language and visual representations, resulting in sub-optimal multimodal representations and decreased downstream task performance. To prevent this from happening, one of the prevalent approaches is to utilize a residual connection, which can improve gradient flow \cite{huang2016deep,huang2017densely,he2016deep} and reinforce a prior representation. However, residual connections contain no information bottleneck, resulting in visual and language representations that contain unrelated information for downstream tasks. From this motivation, we design a multimodal guided residual model, {\pa}, to reinforce salient multimodal representations for downstream tasks (Fig.~\ref{fig:rerep_model_diagram}).

\textbf{Visual-Language Representations: } Similar to existing models \cite{alayrac2022flamingo,li2022blip,li2023blip,zhai2022lit,kim2021vilt}, we first extract visual and language representations using a frozen pre-trained encoder. We used state-of-the-art VL models to extract visual ($V \in \mathbb{R}^{D_{V}}$) and language $L \in \mathbb{R}^{D_{L}}$ representations, such as CLIP \cite{clip}, DualEncoder \cite{wu2019dual}, ViLT \cite{kim2021vilt}, and BLIP-2 \cite{li2023blip}. Here, $D_{V}$ and $D_{L}$ are the dimensions of visual and language representations from the pre-trained encoders. 

\textbf{Multimodal Guided Residual Model: }
We introduce a multimodal guided residual module to reinforce salient portions of modality-specific representations, serving as an information bottleneck over vanilla residual connection \cite{he2016deep} reinforcing entire representations. This is done by focusing on the most relevant parts of the visual or language representations using cross-attention. Cross-attention is similar to self-attention but has a crucial difference in its inputs. In cross-attention, the query is different from the keys and values, whereas in self-attention these are the same. This allows for the usage of projected visual ($V^p$) and language ($L^p$) representations as the query $(q)$, and usage of the originally extracted visual $(V)$ and language $(L)$ representations as the key ($k$) and value ($v$): 

\noindent\resizebox{\linewidth}{!}{
$\{V^g, L^g\} = \text{Cross-Attention}(q=\{V^p,L^p\},\ k=\{V,L\},\ v=\{V,L\})$
}


This design allows for maintaining beneficial aspects of residual connections, such as improved gradient flow and reinforcement of prior representations, while establishing an information bottleneck on the residual connection. After extracting the guided residual representations, they are added to the projected representations as in vanilla residual connections: $V^f, L^f = V^p + V^g, L^p + L^g$. Finally, we fused these representations ($V^f, L^f$) for downstream task learning.




\textbf{Training Model: } To demonstrate the {\pa} model's effectiveness at improving representations, we train for two downstream tasks: comprehending embodied referring expressions designed as an object bounding box prediction and visual-question answering designed as a multiple choice question-answering task. We used a regression loss for the object bounding box prediction task and a classification loss for the multiple-choice question-answering task.

We developed all models using the PyTorch and PyTorch-Lightning deep learning frameworks. We also used the HuggingFace library for pre-trained models (ViLT, Dual Encoder, CLIP, and BLIP-2). We used an embedding size of $512$ for the Dual-Encoder and CLIP models, $768$ for the ViLT model, and $1408$ for the BLIP-2 model. We trained models using the AdamW optimizer with a weight decay regularization set to $0.01$ \cite{adamw} and cosine annealing warm restarts with a cycle length ($T_0$): $\{2,4,6\}$, and cycle multiplier ($T_{mult}$): $2$.  For the Dual Encoder, CLIP, ViLT, and BLIP-2 models doing detection we used a learning rate of $3e^{-5}$, $3e^{-6}$, $3e^{-5}$, and $3e^{-6}$ respectively, and all models for VQA used a learning rate of $1e^{-5}$. We used a batch size of $32$ for all models except BLIP-2 where we used a batch size of $2$ due to the model being much larger. All models for detection were trained for $10$ epochs on {\ds} and $25$ epochs on CAESAR-PRO with a random seed of $33$; and all models for VQA were trained for $20$ epochs with a random seed of $42$.

%% file: sections/experiment_results.tex
\begin{table*}[!t]
\centering
\small
\caption{Comparison of VL models' performance on the E-REF comprehension task, designed as bounding box detection. The results suggest that our multimodal guided residual module, {\pa}, enhances the performance of most baseline multimodal models on the {\ds} and CAESAR-PRO datasets. Best performance numbers in \textbf{bold} face. (V: Visual, L: Language)}
\vspace{-0.1cm}

\begin{tabular}{l|cc|cc|cc|cc|cc}
\multicolumn{11}{c}{\textbf{{\ds} Dataset}} \\
\toprule
\multirow{2}{*}{Models} & \multicolumn{2}{c|}{Without Residual} & \multicolumn{2}{c|}{Vanilla Residual} & \multicolumn{2}{c|}{\pa (V)} & \multicolumn{2}{c|}{\pa (L)} & \multicolumn{2}{c}{\pa (V+L)} \\
 & IOU-25 & IOU-50 & IOU-25 & IOU-50 & IOU-25 & IOU-50 & IOU-25 & IOU-50 & IOU-25 & IOU-50 \\
\midrule
CLIP & 25.80 & 7.67 & 27.22 & 8.35 & \textbf{29.20} & \textbf{9.15} & 28.30 & 7.50 & 26.65 & 7.27 \\
ViLT  & 36.53 & 14.03 & 35.34 & 14.37 & - & - & - & - & \textbf{37.05} & \textbf{14.66} \\
BLIP-2 & \textbf{29.42} & 7.54 & 27.66 & 7.31 & 25.45 & 7.71 & 26.81 & \textbf{7.94} & 16.44 & 3.80 \\
Dual-Encoder & 31.08 & 9.83 & 30.17 & 8.98 & \textbf{31.36} & 8.92 & 29.43 & 9.03 & 31.08 & \textbf{10.68} \\
\midrule
\multicolumn{11}{c}{\textbf{CAESAR-PRO Dataset \cite{patron}}} \\
\midrule
\multirow{2}{*}{Models} & \multicolumn{2}{c|}{Without Residual} & \multicolumn{2}{c|}{Vanilla Residual} & \multicolumn{2}{c|}{\pa (V)} & \multicolumn{2}{c|}{\pa (L)} & \multicolumn{2}{c}{\pa (V+L)} \\ 
 & IOU-25 & IOU-50 & IOU-25 & IOU-50 & IOU-25 & IOU-50 & IOU-25 & IOU-50 & IOU-25 & IOU-50 \\
\midrule
CLIP & 37.92 & 9.82 & 39.43 & 10.83 & \textbf{42.91} & \textbf{11.91} & 39.56 & 10.85 & 39.06 & 10.46 \\
ViLT & 27.96 & \textbf{8.73} & 25.67 & 8.06 & - & - & - & - & \textbf{28.52} & 8.04 \\
Dual-Encoder & 42.52 & \textbf{12.14} & \textbf{42.61} & 11.61 & 36.72 & 8.51 & 37.97 & 10.32 & 37.72 & 11.50 \\
\bottomrule
\end{tabular}%

\label{tab:vl_models_erfe}
\vspace{-0.1in}
\end{table*}

\section{Experimental Setup}

We have incorporated our proposed guided residual module {\pa} into the existing state-of-the-art multimodal models, including CLIP \cite{clip}, DualEncoder \cite{wu2019dual}, ViLT \cite{kim2021vilt}, BLIP-2 \cite{li2023blip}, and VisualBERT \cite{li2019visualbert}. We have evaluated these models and baselines multimodal models on {\ds} and CAESAR-PRO \cite{patron} datasets focusing on embodied referring expression comprehension ({\erfe}) tasks. We have also evaluated these models on two more widely used datasets, ScienceQA \cite{lu2022learn}, and A-OKVQA \cite{schwenk2022okvqa}, to assess their performance on Visual Question Answering (VQA) tasks. We trained multiple variations of our proposed residual module {\pa}, each differing in the type of residual representation of visual and language modalities. We examined four distinct variations:

\begin{itemize}
\item \textbf{Visual-Only Residual Representation {\pa}(V)}: This variant leverages the projected visual representation as the query in the guided residual modules to extract the salient multimodal residual representations.

\item \textbf{Language-Only Residual Representation {\pa}(L)}: This variant utilizes the projected language representation as the query in the guided residual modules to extract the salient multimodal residual representations.

\item \textbf{Visual and Language Residual Representation {\pa} (V+L)}: This variant employs projected visual and language representations as the query to extract the salient multimodal residual representations.

\item \textbf{Vanilla Models}: Following the original residual architecture \cite{he2016deep}, this baseline directly summed visual and language representations to the projected representations without using any attention approach. We also evaluated several multimodal models in the vanilla mode without any residual connections. 
\end{itemize}

\section{Experimental Results}
\subsection{Embodied Referring Expression}

We evaluated models on the {\ds} and CAESAR-PRO datasets for the embodied referring expression comprehension task. Following prior work on the embodied referring expression task \cite{chen2021yourefit}, we designed this task as an object bounding box detection task. All models were trained following a similar setup outlined in Section~\ref{sec:model}. We have reported Top-1 accuracy for the VQA tasks. The experimental results are presented in Table~\ref{tab:vl_models_erfe}.

\textbf{Results and Discussion: } The experimental results in Table~\ref{tab:vl_models_erfe} indicate that augmenting existing multimodal models with the proposed multimodal guided residual module {\pa} enhances embodied referring expression comprehension task performance on both the {\ds} and CAESAR-PRO datasets. More specifically, the results indicate that including \textbf{visual} reinforced representations enhances task performance. For example, augmenting {\pa} into the CLIP\cite{clip} model and reinforcing the visual representation improved the object bounding detection task performance on our {\ds} dataset from $25.80\%$ to $29.20\%$ for IOU-25. Similarly, {\pa} helps the CLIP\cite{clip} model enhance object bounding detection task performance on the CAESAR-PRO \cite{patron} dataset from $37.92\%$ to $42.91\%$ for IOU-25. This performance improvement underscores the importance of visual cues in object grounding and suggests that reinforcing visual representation can lead to better performance. 

Although the vanilla residual connection offers some performance improvement over models without any residual connection-based fusion, the gains are modest compared to those achieved with {\pa}. The key distinction lies in {\pa}'s selective reinforcement of the most salient aspects of the visual-language representation, acting as an information bottleneck to extract only the relevant information. This targeted approach contrasts with vanilla residual connections, which indiscriminately reinforce the entire representation. These insights align with the findings from prior works on the information bottleneck \cite{islam2023representation,wang2022rethinking,tishby2015deep,shwartz2017opening,sun2022graph,alemi2016deep,discrete_bottleneck,islam2024eqamx}. In the literature, it has been shown that information bottleneck helps the model to extract the relevant information and thus improve downstream task performance. Thus, the design choice of residual representation incorporation is pivotal in refining multimodal representation and, consequently, downstream task performance.

The experimental results further suggest that the specific modality being reinforced can influence performance improvements. For example,  reinforcing the visual modality with {\pa} boosts the CLIP model's performance for the object bounding box detection task from $25.80\%$ to $29.20\%$ for IOU-25. Conversely, emphasizing the language modality results in a slightly lower enhancement, with performance increasing to $28.30\%$. This variance suggests that the object grounding task is predominantly reliant on visual information. Thus, the choice of modality for reinforcement should be carefully considered based on the downstream task.

\vspace{-0.05in}
\subsection{Visual Question-Answering Tasks}





We evaluate the models on the ScienceQA \cite{lu2022learn} and A-OKVQA \cite{schwenk2022okvqa} datasets for the VQA task. Visual question answering is central to HRI, as robots must interpret multimodal queries, ground them in visual scenes, and provide accurate responses. Using ScienceQA and A-OKVQA enables us to assess whether our model facilitates such interactive understanding, which is crucial for natural human–robot collaboration. Following the evaluation protocols of these benchmarks, we conduct multiple-choice QA evaluations. Similar to the previous tasks, we incorporate different variations of our multimodal guided residual module {\pa} into CLIP, ViLT, and VisualBERT models: {\pa}(V), {\pa}(L), {\pa}(V+L), and Vanilla multimodal models without residual connections for multimodal fusion. Since ViLT is a monolithic model that provides joint vision–language representations, we split its output into separate text and image representations based on the text length derived from the attention mask. All models were trained following the setup outlined in Section~\ref{sec:model} (Training Model). We report Accuracy for ScienceQA and the Multiple Choice (MC) evaluation metric \cite{schwenk2022okvqa} for A-OKVQA. The experimental results are presented in Table~\ref{tab:vl_models_vqa}.

\textbf{Results and Discussion: } The experimental results in Table~\ref{tab:vl_models_vqa} suggest that incorporating our multimodal guided residual module, {\pa}, into multimodal models demonstrates consistent performance improvement across all variations evaluated compared to those without residual connections. Specifically, the inclusion of both visual and linguistic modalities ({\pa}(V+L)) consistently yields the highest improvements. For example, in the ScienceQA dataset, CLIP model with {\pa} VQA task accuracy increases from $21.31\%$ to $51.85\%$. This performance improvement attributed to the information bottleneck in {\pa} effectively extracts the salient representation from visual and language modalities, leading to more accurate answers. 

\begin{table*}[!t]
\footnotesize
\centering
\caption{Comparison of VL models' performance on the visual question-answering task. The results suggest that our multimodal guided residual module, {\pa}, enhances the performance of the multimodal models on the ScienceQA and A-OKVQA datasets. Best performance numbers in \textbf{bold} face. (V: Visual, L: Language)}
\vspace{-.1cm}
\setlength{\tabcolsep}{4pt}
\begin{tabular}{l|ccccc|ccccc}
\toprule
& \multicolumn{5}{c|}{\textbf{ScienceQA Dataset \cite{lu2022learn}}} & \multicolumn{5}{c}{\textbf{A-OKVQA Dataset \cite{schwenk2022okvqa}}} \\
\midrule
Models & Without$\;$Residual & With$\;$Residual & \pa (V) & \pa (L) & \pa (V+L) & Without$\;$Residual & With$\;$Residual & \pa (V) & \pa (L) & \pa (V+L) \\ 
\midrule
CLIP & 21.31 & 33.36 & 40.75 & 31.33 & \textbf{51.85} & 29.41 & \textbf{32.78} & \textbf{32.78} & 30.42 & 32.47\\
ViLT & 44.52 & 47.05 & 42.78 & 42.58 & \textbf{49.33} & 31.61 & 31.21 & 32.19 & 31.48 & \textbf{32.53}\\
VisualBERT & 34.95 & 36.63 & 37.13 & 37.63 & \textbf{39.03} & 29.88 & 32.47 & 30.72 & 31.15 & \textbf{32.62}\\ 
Dual-Encoder & 24.79 & 35.55 & 37.13 & 31.93 & \textbf{43.57} & 32.64 & 33.45 & 32.89 & 31.72 & \textbf{35.02}\\ 
\bottomrule
\end{tabular}
\label{tab:vl_models_vqa}
\end{table*}

The gains from visual-only ({\pa} (V)) and language-only ({\pa} (L)) reinforcements underscore the importance of modality-specific enhancements, with visual reinforcements being particularly impactful in the VisualBERT model on the ScienceQA dataset, improved its performance from $34.95\%$ to $37.13\%$ using visual reinforcement and $37.63\%$ using language reinforcement. These insights suggest that strategically leveraging multimodal guided residuals can significantly refine model performance in VQA tasks. 

%% file: sections/overal_discussion.tex
\section{Overal Discussion} 
The experiments provide several key insights into the comprehension of E-RFEs. The proposed Refer360 dataset addresses critical gaps in prior work by capturing a broad spectrum of real-world objects and settings for embodied instructions. Unlike earlier datasets constrained to a single view or limited modalities, Refer360 provides synchronized multi-view recordings (egocentric, exocentric, and top-down) of referential behaviors, paired with both verbal utterances and rich nonverbal gestures. Spanning indoor and outdoor environments with varied lighting and context introduces essential environmental diversity. Multi-perspective data enable models to develop perspective-invariant understanding, such as distinguishing ``the object on my left” versus ``your left”—a nuance often missed in single-view settings. Similarly, the inclusion of pointing and gaze cues enables models to learn how these signals jointly disambiguate references, which is crucial for robust comprehension of situated instruction. Beyond serving as a benchmark, Refer360 provides a foundation for fine-tuning large vision–language models (VLMs), grounding abstract multimodal knowledge in embodied contexts.

Alongside the dataset, the MuRes offers a lightweight yet effective fusion mechanism for E-RFEs. Across all evaluation setups, integrating MuRes consistently improved performance over baseline multimodal models, underscoring the importance of reinforcing modality-specific cues rather than relying solely on alignment-based fusions. By applying an information bottleneck to residual connections, MuRes extracts complementary representations that bridge the gap between generic pre-trained encoders and the fine-grained signals required for E-RFEs. Functioning as an adapter layer, MuRes selectively amplifies salient information before fusion, aligning with emerging strategies that augment frozen vision–language backbones with small, trainable modules. In our experiments (Tables~\ref{tab:vl_models_erfe}, \ref{tab:vl_models_vqa}), this design translated into notable improvements in understanding referring expressions and answering questions in human-centric scenes. These findings resonate with other recent efforts to extend general-purpose VLMs with targeted HRI modules, highlighting MuRes as a compact and general-purpose adapter for embodied interaction.

Looking ahead, the integration of structured HRI datasets such as Refer360 with modular approaches (e.g., MuRes) paves the way for more embodied and intelligent robots. For example, Refer360 can be extended beyond embodied referring expressions to broader visual question-answering (VQA) tasks, opening the possibility of evolving into an embodied question-answering (E-QA) benchmark that integrates situated reasoning with multimodal embodied cues to enable more intuitive HRI. We envision systems that leverage the broad knowledge of foundation models while being grounded through fine-tuning or adapter-based learning on real interaction data. Evidence from hybrid approaches—such as PaLM-E, which combines internet-scale text with robot sensor inputs to achieve state-of-the-art embodied performance—demonstrates the promise of this direction. A lightweight alternative is to begin with capable VLMs and inject HRI expertise through adapter modules, targeted training, or prompt-based alignment. By doing so, future robots could interpret human instructions with the flexibility of large language models while retaining the situational grounding required for natural collaboration. The combined contributions of Refer360 and MuRes provide concrete tools toward this vision, advancing progress toward robots that can naturally comprehend and engage with humans in real-world environments.

%% file: sections/conclusion.tex
\vspace{-.2cm}
\section{Conclusion}

In this paper, we have introduced a comprehensive and diverse benchmark dataset of multimodal interactions, {\ds}, and explored a baseline adapter model, {\pa}, to extract modality-specific salient representations. To comprehensively study embodied referring expressions in real-world settings, we have curated {\ds} from various environments, collecting multimodal sensor data including exo visual view, ego visual view, depth, infrared, 3D skeletal data, audio, and robot camera motion to capture unconstrained human interactions from multiple verbal and visual viewpoints. Consequently, {\ds} is the first embodied referring expression comprehension dataset curated with such diverse sensor data. Our extensive experimental analyses demonstrate that existing multimodal models cannot effectively understand embodied referring expressions in real-world settings, primarily due to their failure to bridge the gap between general pre-trained frozen visual-language representations and salient modality-specific cues. To address this issue, we explored a baseline multimodal guided residual module, {\pa}, which acts as a bottleneck to extract salient modality-specific representations. Our experiments suggest that incorporating {\pa} into existing multimodal models improves downstream task performance on four datasets comprising embodied referring expression understanding and visual question answering. Our comprehensive multimodal benchmark dataset ({\ds}), along with our exploration of {\pa}, show promising directions for research into embodied referring expression comprehension.

%% file: supplementary/resources.tex
\section{Resources}
\raggedright
\textbf{Processed Dataset (49.71 GB):} \\
\url{https://bit.ly/refer360-dataset-processed}

\textbf{Raw Dataset (2572.36 GB):} \\ The link will be provided upon acceptance.

\textbf{Source Code (Data Collection):} \\ The link will be provided upon acceptance.

\textbf{Source Code (\pa{} + baselines):} \\ The link will be provided upon acceptance.

\textbf{Model Checkpoints (5.4 GB):} \\ The link will be provided upon acceptance.

\textbf{Docker for training models (8.59 GB):} \\
We built a Docker to facilitate easy
reproduction of our experimental settings and 
training environment. We cannot currently share the Docker Hub link to maintain anonymity. We plan to share that link upon publication of the paper. For now, we  share the Singularity container built from the same Docker used in our experiments:\\
\url{https://bit.ly/multimodal-docker}

%% file: supplementary/experiments_results.tex
\section{Additional Experimental Results: Quantitative Analysis}

\begin{figure}[!t]
    \centering
    \small
    \begin{tabular}{c}
        \includegraphics[width=0.95\linewidth]{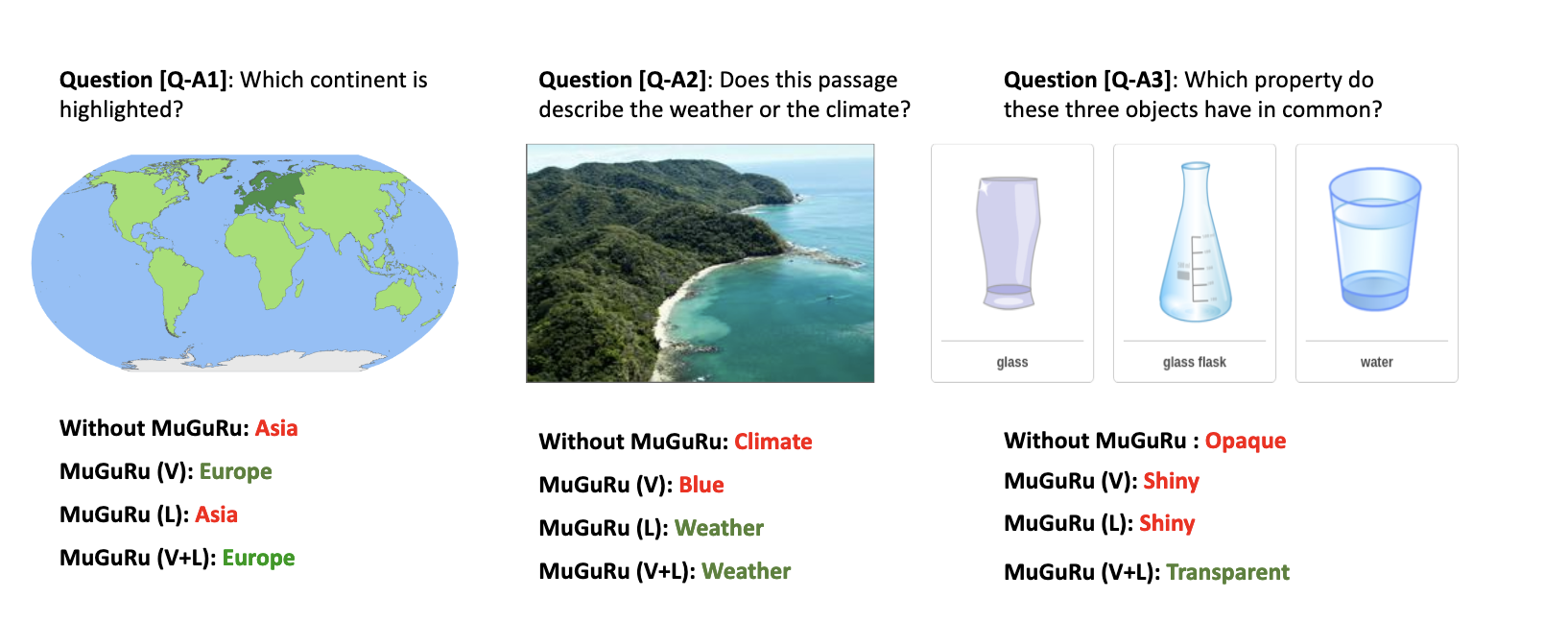} \\
        \textbf{(a)} ScienceQA \cite{lu2022learn} \\
        \includegraphics[width=0.95\linewidth]{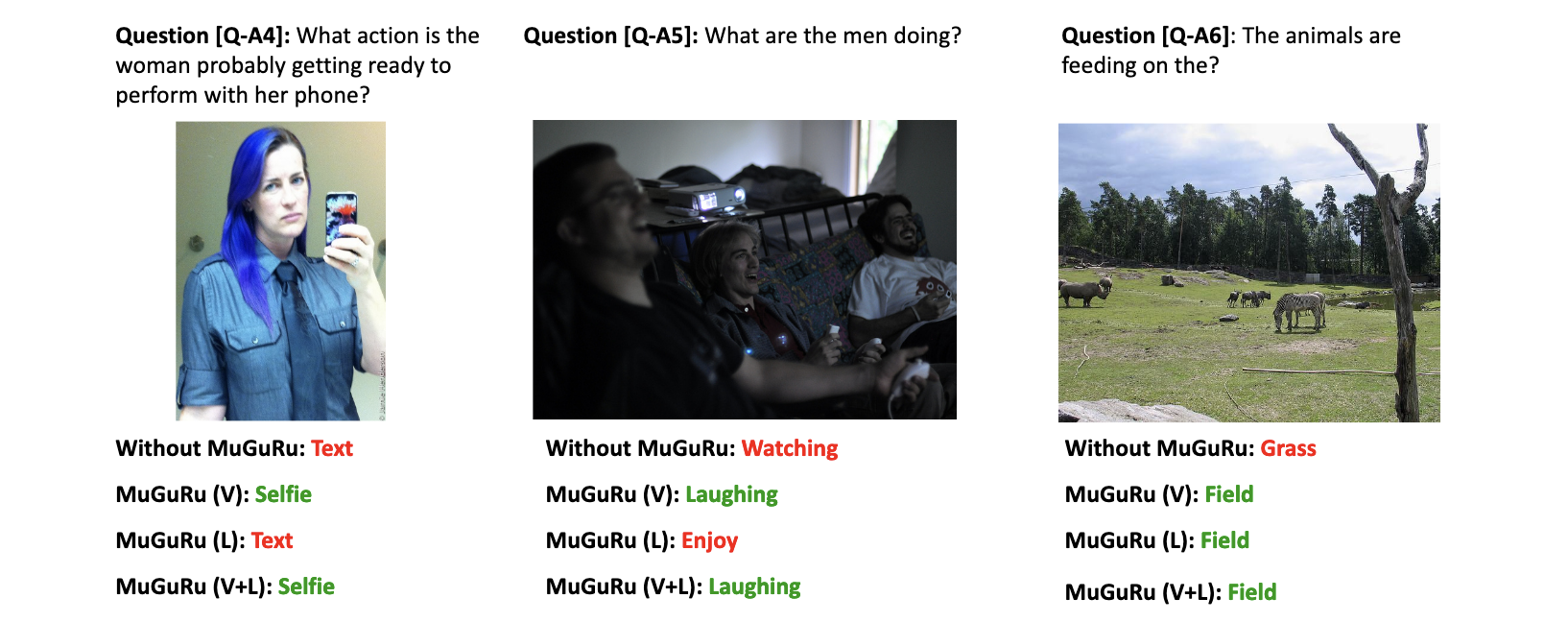} \\
        \textbf{(b)} A-OKVQA \cite{schwenk2022okvqa} \\
    \end{tabular}
    \caption{Qualitative analysis of VisualBERT with and without the proposed guided residual module (\pa) on two datasets. Incorporating {\pa} improves visual question answering by aligning multimodal context in both ScienceQA and A-OKVQA tasks. Subfigure (a) shows qualitative results on questions involving diagrams from the ScienceQA dataset \cite{lu2022learn}. Subfigure (b) shows model reasoning on ambiguous visual questions from the A-OKVQA dataset \cite{schwenk2022okvqa}.}
    \Description{Qualitative analysis of VisualBERT with and without the proposed guided residual module (\pa) on two datasets. Incorporating {\pa} improves visual question answering by aligning multimodal context in both ScienceQA and A-OKVQA tasks. Subfigure (a) shows qualitative results on questions involving diagrams from the ScienceQA dataset \cite{lu2022learn}. Subfigure (b) shows model reasoning on ambiguous visual questions from the A-OKVQA dataset \cite{schwenk2022okvqa}.}
    \label{fig:qualitative}
\end{figure}

We have performed a quantitative evaluation of the models by applying the ScienceQA \cite{lu2022learn} and A-OKVQA \cite{schwenk2022okvqa} datasets for the visual-question answering tasks. We have analyzed the response of VisualBERT with different variations of our proposed model, (\pa), on multiple-choice question-answering tasks. The responses from VisualBERT model variations are similar to the variation presented in Table~\ref{fig:qualitative} from the manuscript (i.e., without residual, {\pa} (V), {\pa} (L), and {\pa} (V+L)).

\textbf{Discussion: } The model responses are presented in Fig.~\ref{fig:qualitative}. These results suggest that augmenting the VisualBERT model with {\pa} improves responses for the visual question-answering task. For instance, in Fig.~\ref{fig:qualitative} (a) [Q-A1], the VisualBERT model's response to the question \textit{``Which continent is highlighted?''} alongside an image of a map shows that enhancing visual representations through {\pa} yields the correct answer (\textit{``Europe''}). However, enhancing only the language representations through {\pa} leads to an incorrect answer (\textit{``Asia''}). This question necessitates a thorough understanding of the spatial location of the highlighted region (\textit{``Europe''}) on the map, explaining why reinforcing the visual representations aids in improving the response. Conversely, in Fig.\ref{fig:qualitative} (a) [Q-A3], enhancing either visual or language representations does not yield the most accurate answer (\textit{``Transparent''}) for the question: \textit{``Which property do these three objects have in common?''}. Although the responses with either Vision or Language in Fig.\ref{fig:qualitative} (a) [Q-A3] are not entirely inaccurate, as the objects are somewhat shiny, only yhe model with both visual and language representations reinforced correctly answers ``Transparent''. Therefore, identifying which modalities to reinforce thorough {\pa} is a critical aspect of enhancing the model's responses.

%% file: supplementary/appendix_data_collection_system.tex
\section{Data Collection}
%
%
\subsection{Data Collection System}
Our data collection system integrates an Azure Kinect DK \cite{azure-kinect} a and a Pupil Smart Glass, also known as the Pupil Invisible Eye Tracker \cite{pupil-labs}. The Azure Kinect  DK was mounted on an Ohmni Telepresence robot \cite{ohmnilabs}, and the participants wore the Pupil Smart Glass to facilitate data collection in real-world scenarios. An Alienware m15 R4 laptop powered by an i7-10870H RTX processor served as the high-performance computing backbone. A Python-based application was developed to facilitate coordination and synchronization among all system components. This application ensured seamless operation and synchronized data collection from multiple sensors.
\subsubsection{Sensor Specifications}

 Azure Kinect provides a multitude of sensory data, including visual, depth, infrared (IR), skeletal tracking, and inertial measurement unit (IMU) data. In addition, pupil glass offers visual (RGB), IR, gaze tracking, and gesture recognition capabilities.  The Pupil Invisible Eye Tracker is a state-of-the-art device with a range of features designed to capture precise and accurate eye-tracking data. The participants in our study were equipped with the Pupil Smart Glass and an Android smartphone, which recorded their eye-tracking data. The data is subsequently transmitted to the Pupil Cloud via the Pupil Invisible Android application. This seamless hardware and software integration ensures efficient and reliable data collection and transmission. The specifications of the Azure Kinect DK  and Pupil Eye Tracker sensors are listed in Table \ref{tab:azure_kinect_specs} and \ref{tab:pupil_specs}.

%

\
\subsubsection{Data Collection Application}

We developed a Python application to coordinate and synchronize the various components of our data collection system. This application played a central role, ensuring seamless integration and synchronized data capture from multiple sensors. We collected camera video feeds, time-series data from the inertial measurement unit (IMU) and skeleton joint positions, and session metadata using this system. We utilized the pyKinectAzure \cite{pyKinectAzure} python library to interface with the Azure Kinect SDK sensor, while the Pupil Labs' Real-time Python API \cite{pupil-labs-realtime-api} facilitated communication with the Pupil Eye camera. Participants stood before the Ohmin robot, issuing verbal commands and non-verbal gestures to reference physical objects. An RGB camera on the Azure Kinect device continuously captured visual data, providing a third-person view of the participants' referencing gestures.
Additionally, the Kinect's depth and infrared sensors recorded supplementary data streams, enriching the external perspective of the interactions. The system also leveraged the Kinect's infrared sensor to collect infrared data and the Azure Kinect Body Tracking SDK \cite{azure-kinect-body-tracking} to capture the 3D coordinates and orientations of 32 skeletal joints. Simultaneously, the Kinect's microphone recorded the participants' verbal instructions. Complementing this external viewpoint, the Pupil Invisible Eye Tracker provided an egocentric visual stream from the participants' perspectives. Combining these exocentric and egocentric data sources gave the system a comprehensive understanding of human-robot interactions.

We stored the Azure Kinect recordings and the corresponding keystroke event times locally as MP4 and JSON files, respectively. For the Pupil eye tracker, the recordings of the participants' ego view and keystroke events were saved in the Pupil Cloud using the Pupil Lab Android app and Pupil API, respectively.

\subsubsection{Time-based Synchronization}
One of the significant challenges we faced was synchronizing the various data streams captured by different devices. To address this, we implemented a time-based synchronization method that recorded the UNIX timestamps of different data capture events and data streams, enabling synchronization during post-processing. This synchronization is crucial for aligning the data streams captured from different devices.
Our approach involved recording the timestamp at the start and end of each interaction and the timestamp of the event when the participant pointed to an object (i.e., canonical events). This was achieved using our Python-based system, which is operated by individuals recording the data collection sessions. We utilized different keystrokes on a standard keyboard to denote different events. The ``Space" key was pressed at the start and end of an interaction, while the ``G" key was pressed to identify the canonical event of an interaction. The canonical event indicates when the participant points to an object using gaze or pointing gestures. Specifically, the ``G" keystroke event time was used to identify the canonical frame, i.e., the frame where the participant actually pointed to an object. When the participant used cues other than pointing, such as gaze, the ``G" key was pressed when the gaze event occurred. The ``Space" keystroke event time was used to identify the start and end of an interaction, thereby facilitating the segmentation of interactions. The ``Q" key was used to terminate a session.

\begin{figure}[!t]
    \centering
    \fbox{%
        \begin{minipage}{0.95\linewidth}
            \scriptsize
            grey ceramic bowl, foam miniature football, wireless computer mouse, wooden box, blue cupholders, plastic water bottle, keyboard, green plastic cup, white plastic basket, basketball, white plastic cup, flower vase, clorox wipe container, paper towel roll, mountain dew bottle, picture frame, TV remote, grey plastic basket, black metal water bottle, coffee cup with lid, transformers robot, pepsi bottle, egg carton, TV screen, blue plastic box, pringles box, grey dustbin, light green open plastic box with handle, tripod, white three-level plastic box, cardboard box, sunglasses, yellow lego box, mouthwash, pink plastic cup, white tumbler, white desk fan, blue plastic container with lid and handle, salsa jar, nutella jar, pink dustbin, black kickball, table tennis ball container, blue plastic water bottle, black desk clock, screwdriver, blue magazine, shoe rack, bicycle, pupil labs glasses box, microwave, frying pan, blue couch, wooden chair, white rope, kitchen sink, white fridge, iron stand, allen wrench set, white trash can, black dresser, light stand, desk lamp, black office chair, silver rice cooker, black standing fan, wooden table, white pillow, white air conditioning unit, grey sweatshirt, banana, grey laundry drying rack, grey apartment mailboxes, white fence, surge protector.
        \end{minipage}%
    }
    \caption{Objects in Refer360 Dataset}
    \Description{Objects in Refer360 Dataset}
    \label{fig:object-list}
\end{figure}

The corresponding UNIX timestamp for these keystroke events was recorded for both the Azure Kinect and Pupil Lab Eye Tracker. This enabled us to synchronize the data streams from these two devices during post-processing. Though the time-based synchronization method is utilized to synchronize between the Azure Kinect Sensor and Pupil Eye Tracker, it is designed to be extensible. For example, our system can be expanded to incorporate multiple Azure Kinect devices to capture multiple views of the participant during interaction rather than just the ego and exo views.

\begin{table}[!t]
\centering
\small
\caption{Azure Kinect DK Sensor Specifications}
\label{tab:azure_kinect_specs}
\begin{tabular}{@{}p{3.2cm}p{4.9cm}@{}}
\toprule
\textbf{Sensor}      & \textbf{Specification} \\ \midrule
RGB Camera           & Resolution: $3840 \times 2160$ @ 30 fps \\
Depth Camera         & Time-of-Flight; Resolution: $640 \times 576$ @ 30 fps \\
Motion Sensor        & LSM6DSMUS IMU (accel. \& gyro), Sampling Rate: 1.6 Hz \\
Microphone Array     & USB Audio 2.0; 7 channels; Sensitivity: $-22$ dBFS (94 dB SPL, 1 kHz); SNR: $>65$ dB; Overload Point: 116 dB \\
\bottomrule
\end{tabular}
\end{table}

\begin{table}[!t]
\centering
\small
\caption{Pupil Invisible Eye Tracker Specifications}
\label{tab:pupil_specs}
\begin{tabular}{@{}p{3.2cm}p{4.9cm}@{}}
\toprule
\textbf{Sensor}     & \textbf{Specification} \\ \midrule
Eye Cameras         & $200$ Hz @ $192 \times 192$ px, infrared (IR) illumination \\
Scene Camera        & $30$ Hz @ $1088 \times 1080$ px, $82^\circ \times 82^\circ$ field of view (FOV) \\
\bottomrule
\end{tabular}
\end{table}

\subsubsection{Data Collection Environment} 
The Refer360 dataset aims to study real-world human-robot interactions in which a human provides object-referencing instructions to robots across diverse environments, ranging from controlled laboratory setups to outdoor locations. Refer360 contains embodied interaction data from lab and outside-lab environments. The outside lab refers to settings outside controlled lab settings, such as homes, outdoor locations, etc. While choosing objects,
we prioritize those usually available in these environments. Our dataset contains 75 objects from the aforementioned environments, and a complete list of objects is given in Fig.~\ref{fig:object-list}.


%
%

\begin{figure}[!t]
    \centering
    \includegraphics[width=0.85\columnwidth]{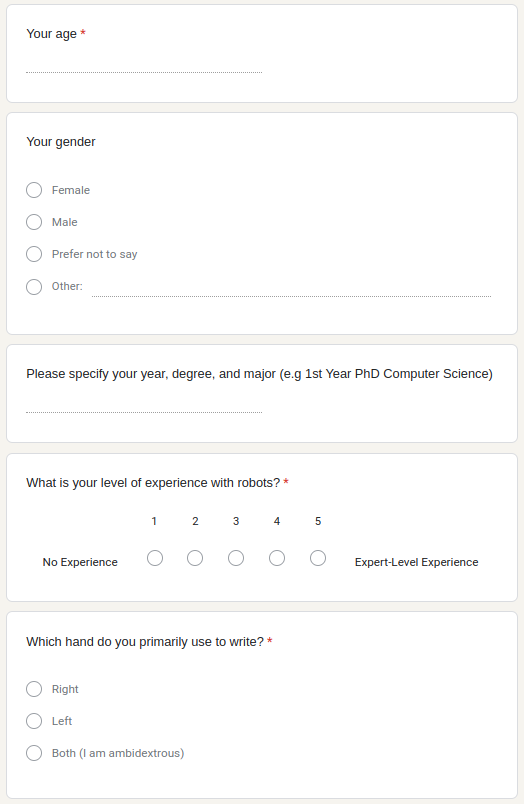}
    
    \vspace{1mm}
    \textbf{(a)} Demographic Survey
    
    \vspace{3mm}
    \includegraphics[width=0.85\columnwidth]{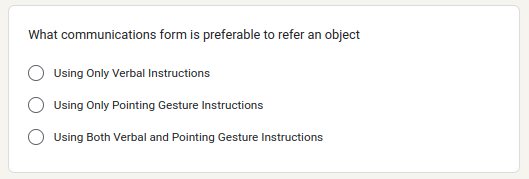}
    
    \vspace{1mm}
    \textbf{(b)} Post-Task Survey

    \caption{User interfaces shown during data collection. (a) Participants first completed a demographic survey. (b) After viewing the stories, participants completed a post-task evaluation survey.}
    \Description{User interfaces shown during data collection. (a) Participants first completed a demographic survey. (b) After viewing the stories, participants completed a post-task evaluation survey.}
    \label{fig:survey_interfaces}
\end{figure}

The data collection process began with a comprehensive introduction to the system, the purpose of the dataset, and the protocol to be followed during collection. Before participating in the data collection sessions, subjects completed a demographic survey.

Each session involved subjects providing embodied instructions that referenced objects in their surroundings, using both language and nonverbal gestures (gaze and pointing gestures). The ultimate goal of this dataset is to enhance social robots' ability to interpret object referencing instructions accurately. This involves uniquely identifying the object, which requires extracting the object's location and other attributes from the instruction. This task is challenging as humans often use diverse formats when providing verbal instructions, and these instructions may sometimes lack the necessary features for object identification. Incorporating nonverbal cues, such as pointing or referencing the object in relation to another object, can significantly improve the efficiency of interpreting object referencing instructions. Furthermore, object referencing instructions can be given from multiple perspectives, such as the subject's or the robot's perspective, which must be resolved for accurate object comprehension.

 The participants were given the flexibility to choose any perspective (subject, robot, or neutral) when providing instructions. This approach allowed us to diversify our dataset by including object-referencing instructions with varied spatial referencing and perspectives. For instance, an object could be referenced in relation to another object, such as ``The black box on top of the brown table." The object reference in the verbal instruction could be from the subject's perspective, e.g., ``The couch to my right," or it could be from the robot's perspective, e.g., ``The lamp to your left."

We had two distinct data collection conditions: constrained and unconstrained. In the constrained condition, subjects were briefed on the format of instructions and how they could employ various modalities (verbal and nonverbal) to make the interaction as natural as possible. We also suggested that participants use both verbal and nonverbal gestures to describe an object. In the unconstrained condition, we did not suggest whether to use verbal or nonverbal gestures to describe an object. We instructed the participant to describe an object to the robot. This allowed us to capture natural human instincts when providing instructions. This approach also helped eliminate biases that might be introduced by pre-guidance on the format of the instructions, allowing subjects to be flexible in their instruction delivery.

Each subject participated in multiple sessions, each lasting approximately one hour. During each session, the subject performed several interactions. Using our data collection system, we recorded the subject's ego view, exo view, IMU, skeleton, and audio data stream for each session. Upon completion of the sessions, subjects were asked to complete a post-task survey and sign a consent form to permit the release of the dataset. The University's IRB approved the study. The demographic and post-task surveys are presented in Figure \ref{fig:survey_interfaces}.

\begin{figure}[!t]
    \centering
    \includegraphics[width=0.65\columnwidth]{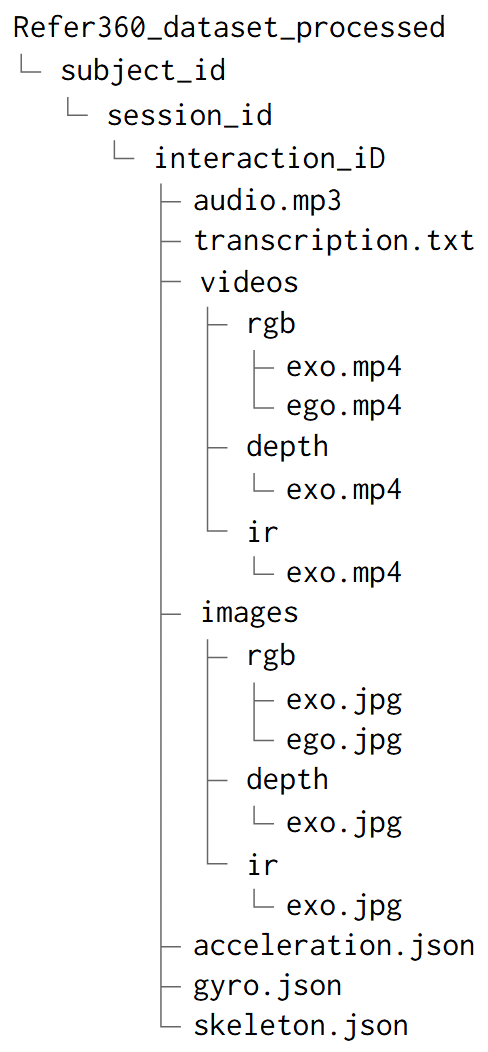}

    \caption{Folder structure of the Refer360 dataset. Each interaction contains audio, video, skeletal, and sensor data across RGB, depth, and infrared modalities.}
    \label{fig:folder_structure}
\end{figure}

\subsection{Dataset Processing}
The developed Python-based application generated an Azure Kinect video file in MP4 format for each session. The MP4 file contains three data streams from Azure Kinect's camera sensor: RGB, Depth, and Infrared. Separate JSON files contain the IMU and skeleton joints' time series data and relevant session metadata. We utilized the FFmpeg  \cite{FFmpeg}  library to extract the Kinect video streams into separate MP4 files and the recording audio as an MP3 file. The IMU time series was split into two different files for the accelerometer and gyroscope readings. For each session, the Pupil eye tracker also generated one video file in MP4 format and saved it to the pupil cloud.

The major challenge of data post-processing was segmenting the interactions and synchronizing the Azure Kinect and Pupil lab data. For the segmentation of each interaction from Azure Kinect data streams, we look into that interaction's start and end time. We also identify the canonical frames, i.e., frames where the subject points precisely to the object. We split each interaction and canonical frame using the FFmpeg library. Next, we searched the corresponding Pupil recording for the Azure Kinect recording from the pupil cloud using Python Pupil Cloud API. For this purpose, we used the recording-start timestamp saved in the metadata file to find the matching Pupil recording in the pupil cloud. After downloading the Pupil video, we employed the same procedure as Azure Kinect recording to split the interactions and canonical frames at the timestamps recorded during data collection. Finally, we utilized the OpenAI whisper \cite{openai-whisper} library to transcribe Kinect audio data to the corresponding text. Note that we manually verified the synchronization and segmentation with five human experts whom the IRB approved. Subsequently, the dataset underwent annotation by human annotators sourced from an external company specializing in data annotation services, ensuring accuracy and reliability.

Our dataset contains several data collection sessions and after data post-processing results in each session's folder structure shown in Figure \ref{fig:folder_structure}. Here, \textit{transcription.txt} is the text transcription of {audio.mp3}. In the subfolders in \textit{Videos} and \textit{Frames}s, \textit{exo.mp4} and \textit{ego.mp4} refer to the videos from the Azure Kinect SDK camera and Pupil Eye Camera, respectively.